\documentclass[11pt,a4paper]{article}
\usepackage[hyperref]{naaclhlt2018}
\usepackage{latexsym}

\usepackage{url}

\aclfinalcopy 
\def\aclpaperid{143} 

\usepackage{mathptmx}
\usepackage{comment}
\usepackage{enumitem}
\setlist{nolistsep}
\usepackage{comment}
\usepackage{pifont} 
\usepackage{graphicx}

\usepackage{booktabs,tabularx}
\newcolumntype{Y}{>{\raggedright\arraybackslash}X} 

\newif\ifarxiv
\arxivtrue

\title{The Argument Reasoning Comprehension Task:\\Identification and Reconstruction of Implicit Warrants}

\author{
Ivan Habernal$^{\dagger}$ \quad Henning Wachsmuth$^{\ddagger}$ \quad Iryna Gurevych$^{\dagger}$ \quad Benno Stein$^{\ddagger}$ \\[.3em]
$^\dagger$ Ubiquitous Knowledge Processing Lab (UKP) and Research Training Group AIPHES  \\
Department of Computer Science, Technische Universit\"{a}t Darmstadt, Germany\\
{\tt www.ukp.tu-darmstadt.de} \quad {\tt www.aiphes.tu-darmstadt.de} \\
$^{\ddagger}$ Faculty of Media, Bauhaus-Universit\"{a}t Weimar, Germany \\
{\tt <firstname>.<lastname>@uni-weimar.de}
}

\date{}

\begin{document}

\ifarxiv
\onecolumn
\noindent \textbf{The Argument Reasoning Comprehension Task: Identification and Reconstruction of Implicit Warrants}

\medskip
\noindent Ivan Habernal, Henning Wachsmuth, Iryna Gurevych, Benno Stein

\bigskip
This is a \textbf{pre-print non-final version} of the article accepted for publication at the \emph{2018 Conference of the North American Chapter of the Association for Computational Linguistics: Human Language Technologies (NAACL 2018)}. The final official version along with the supplementary materials will be published on the ACL Anthology website in June 2018: \url{http://aclweb.org/anthology/}

\medskip
Please cite this pre-print version as follows.
\medskip

\begin{verbatim}
@InProceedings{habernal.et.al.2018.NAACL.arct,
  title = {The Argument Reasoning Comprehension Task: Identification
           and Reconstruction of Implicit Warrants},
  author = {Habernal, Ivan and Wachsmuth, Henning and
            Gurevych, Iryna and Stein, Benno},
  publisher = {Association for Computational Linguistics},
  booktitle = {Proceedings of the 2018 Conference of the North American
               Chapter of the Association for Computational Linguistics:
               Human Language Technologies},
  pages = {(to appear)},
  month = jun,
  year = {2018},
  address = {New Orleans, LA, USA}
  url = {https://arxiv.org/abs/1708.01425}
}
\end{verbatim}
\twocolumn
\fi

\maketitle
\begin{abstract}
Reasoning is a crucial part of natural language argumentation. To comprehend an argument, one must analyze its {\em warrant}, which explains why its claim follows from its premises. As arguments are highly contextualized, warrants are usually presupposed and left implicit. Thus, the comprehension does not only require language understanding and logic skills, but also depends on common sense. In this paper we develop a methodology for reconstructing warrants systematically. We operationalize it in a scalable crowdsourcing process, resulting in a freely licensed dataset with warrants for 2k authentic arguments from news comments.\footnote{Available at  \url{https://github.com/UKPLab/argumentreasoning-comprehension-task/}, including source codes and supplementary materials.} On this basis, we present a new challenging task, the \emph{argument reasoning comprehension task}. Given an argument with a claim and a premise, the goal is to choose the correct implicit warrant from two options. Both warrants are plausible and lexically close, but lead to contradicting claims. A solution to this task will define a substantial step towards automatic warrant reconstruction. However, experiments with several neural attention and language models reveal that current approaches do not suffice.
\end{abstract}

\section{Introduction}

\emph{Most house cats face enemies}. \emph{Russia has the opposite objectives of the US}. \emph{There is much innovation in 3-d printing and it is sustainable}. 

\smallskip
What do the three propositions have in common? They were never uttered but solely {presupposed} in arguments made by the participants of online discussions. Presuppositions are a fundamental pragmatic instrument of natural language argumentation in which parts of arguments are left unstated. This phenomenon is also referred to as common knowledge \cite[p.~218]{Macagno.Walton.2014}, enthymemes \cite[p.~12]{Walton.2007a}, tacit major premises \cite[p.~319]{Amossy.2009}, or implicit {\em warrants} \cite[p.~8]{Newman.1991}. \newcite{Wilson.Sperber.2004} suggest that, when we comprehend arguments, we reconstruct their warrants driven by the cognitive principle of relevance. In other words, we go straight for the interpretation that seems most relevant and logical within the given context \cite{Hobbs.et.al.1993}. Although any incomplete argument can be completed in different ways \cite{Plumer.2016}, it is assumed that certain knowledge is shared between the arguing parties \cite[p.~180]{Macagno.Walton.2014}.

Filling the gap between the claim and premises (aka reasons) of a natural language argument empirically remains an open issue, due to the inherent difficulty of reconstructing the world knowledge and reasoning patterns in arguments. In a direct fashion, \newcite{Boltuzic.Snajder.2016.ArgMinWS} let annotators write down implicit warrants, but they concluded only with a preliminary analysis due to large variance in the responses. In an indirect fashion, implicit warrants correspond to major premises in argumentation schemes; a concept heavily referenced in argumentation theory \cite{Walton.2012}. However, mapping schemes to real-world arguments has turned out difficult even for the author himself.

Our main hypothesis is that, even if there is no limit to the tacit length of the reasoning chain between claims and premises, it is possible to systematically reconstruct a meaningful warrant, depending only on what we take as granted and what needs to be explicit. As warrants encode our current presupposed world knowledge and connect the reason with the claim in a given argument, we expect that other warrants can be found which connect the reason with a different claim. In the extreme case, there may exist an {\em alternative warrant} in which the same reason is connected to the opposite claim.

The intuition of alternative warrants is key to the systematic methodology that we develop in this paper for reconstructing a warrant for the original claim of an argument. In particular, we first~`twist' the stance of a given argument, trying to plausibly explain its reasoning towards the opposite claim. Then, we twist the stance back and use a similar reasoning chain to come up with a warrant for the original argument. As we discuss further below, this works for real-world arguments with a missing piece of information that is taken for granted and considered as common knowledge, yet, would lead to the opposite stance if twisted.

We demonstrate the applicability of our methodology in a large crowdsourcing study. The study results in 1,970 high-quality instances for a new task that we call \emph{argument reasoning comprehension}: Given a reason and a claim, identify the correct warrant from two opposing options. An example is given in Figure \ref{fig:example1}. A solution to this task will represent a substantial step towards automatic warrant reconstruction. However, we present experiments with several neural attention and language models which reveal that current approaches based on the words and phrases in arguments and warrants do not suffice to solve the task.

\begin{figure}[t!]

\small \noindent 
\makebox[\linewidth]{\rule{\linewidth}{0.1pt}}
\textbf{Title:} Is Marijuana a Gateway Drug? 
\textbf{Description:} Does using marijuana lead to the use of more dangerous drugs, making it too dangerous to legalize?

\smallskip
\textbf{Reason:} Milk isn't a gateway drug even though most people drink it as children. 
And since \emph{\{Warrant 1 $|$ Warrant 2\}}, 
\textbf{Claim:} Marijuana is not a gateway drug.

\smallskip
\makebox[1em][l]{\ding{52}} \textbf{Warrant 1:} milk is similar to marijuana

\smallskip
\makebox[1em][l]{\ding{55}} \textbf{Warrant 2:} milk is not marijuana\\[-1ex]
\makebox[\linewidth]{\rule{\linewidth}{0.1pt}}
\vspace{-2.0em}
\caption{\label{fig:example1} Instance of the argument reasoning comprehension task. The correct warrant has to be identified. Notice the fallacious presupposed false analogy used by the author to make the argument.}
\end{figure}

The main contributions of this paper are (1) a \emph{methodology} for obtaining implicit warrants realized by means of scalable crowdsourcing and (2) a new \emph{task} along with a high-quality dataset. In addition, we provide (a) 2,884 user-generated arguments annotated for their stance, covering 50+ controversial topics, (b) 2,026 arguments with annotated reasons supporting the stance, (c) 4,235 rephrased reason gists, useful for argument summarization and  sentence compression, and (d) a method for checking the reliability of crowdworkers in document and span labeling using traditional inter-annotator agreement measures.

\section{Related Work}

It is widely accepted that an argument consists of a \emph{claim} and one or more \emph{premises} (reasons) \cite{Damer.2013}. \newcite{Toulmin.1958} elaborated on a model of argument in which the reason supports the claim on behalf of a \emph{warrant}. The abstract structure of an argument then is \emph{Reason} $\rightarrow$ (since) \emph{Warrant} $\rightarrow$ (therefore) \emph{Claim}. The warrant takes the role of an inference rule, similar to the \emph{major premise} in Walton's terminology \cite{Walton.2007}.

In principle, the chain \emph{Reason} $\rightarrow$ {\em Warrant} $\rightarrow$ {\em Claim} is applicable to deductive arguments and syllogisms, which allows us to validate arguments properly formalized in propositional logic. However, most natural language arguments are in fact inductive \cite[p.~255]{Govier.2010} or defeasible \cite[p.~29]{Walton.2007a}.\footnote{A recent empirical example is provided by \newcite{Walker.et.al.2014.ArgMinWS} who propose possible approaches to identify patterns of inference from premises to claims in vaccine court cases. The authors conclude that it is extremely rare that a reasoning is explicitly laid out in a deductively valid format.} Accordingly, the unsuitability of formal logic for natural language arguments has been discussed by argumentation scholars since the 1950's \cite{Toulmin.1958}. To be clear, we do not claim that arguments cannot be represented logically (e.g., in predicate logic), however the drift to \emph{informal logic} in the 20th century makes a strong case that natural language argumentation is more than modus ponens \cite{vanEemeren.et.al.2014}.

In argumentation theory, the notion of a \emph{warrant} has also been contentious. Some argue that the distinction of warrants from premises is clear only in Toulmin's examples but fails in practice, i.e., it is hard to tell whether the reason of a given argument is a premise or a warrant \cite[p.~205]{vanEemeren.et.al.1987}. However, \newcite{Freeman.2011} provides alternative views on modeling an argument. Given a claim and two or more premises, the argument structure is \emph{linked} if the reasoning step involves the logical conjunction of the premises.
If we treat a warrant as a simple premise, then the linked structure fits the intuition behind Toulmin's model, such that premise and warrant combined give support to the claim. For details, see \cite[Chap.~4]{Freeman.2011}.

What makes comprehending and analyzing arguments hard is that claims and warrants are usually implicit \cite[p.~82]{Freeman.2011}. As they are `taken for granted' by the arguer, the reader has to infer the contextually most relevant content that she believes the arguer intended to use. To this end, the reader relies on common sense knowledge \cite{Oswald.2016,Wilson.Sperber.2004}.

The reconstruction of implicit premises has already been faced in computational approaches. 
In light of the design of their argument diagramming tool, \newcite{Reed.2004} pointed out that the automatic reconstruction is a task that skilled analysts find both taxing and hard to explain. More recently, \newcite{Feng.Hirst.2011} as well as \newcite{Green.2014.ArgMinWS} outlined the reconstruction of missing enthymemes or warrants as future work, but they never approached it since.
To date, the most advanced attempt in this regard is from \newcite{Boltuzic.Snajder.2016.ArgMinWS}. The authors let annotators `reconstruct' several propositions between premises and claims and investigated whether the number of propositions correlates with the semantic distance between the claim and the premises. However, they conclude that the written warrants heavily vary both in depth and in content.
By contrast, we explore cases with a missing single piece of information that is considered as common knowledge, yet leading to the opposite conclusion if twisted.
Recently, \newcite{Becker.et.al.2017.NLDB} also experimented with reconstructing implicit knowledge in short German argumentative essays. In contrast to our work, they used expert annotators who iteratively converged to a single proposition.

As the task we propose involves natural language comprehension, we also review relevant work outside argumentation here. In particular, the goal of the semantic inference task \emph{textual entailment}  is to classify whether a proposition entails or contradicts a hypothesis \cite{Dagan.et.al.2009}. A similar task, \emph{natural language inference}, was boosted by releasing the large SNLI dataset \cite{Bowman.et.al.2015} containing 0.5M entailment pairs crowdsourced by describing pictures. While the understanding of semantic inference is crucial in language comprehension, argumentation also requires coping with phenomena beyond~semantics.
\newcite{Rajpurkar.et.al.2016.EMNLP} presented a large dataset for reading comprehension by answering questions over Wikipedia articles (SQuAD). In an analysis of this dataset \newcite{Sugawara.Aizawa.2016.EMNLP-WS} found, though, that only 6.2\% of the questions require causal reasoning, 1.2\% logical reasoning, and 0\% analogy. In contrast, these reasoning types often make up the core of argumentation \cite{Walton.2007}.
\newcite{Mostafazadeh.et.al.2015.NAACL} introduced the \emph{cloze story test}, in which the appropriate ending of a narrative has to be selected automatically. The overall context of this task is completely different to ours. Moreover, the narratives were written from scratch by explicitly instructing crowd workers, whereas our data come from genuine argumentative comments.
Common-sense reasoning was also approached by \newcite{Angeli.Manning.2014} who targeted the inference of common-sense facts from a large knowledge base. Since their logical formalism builds upon an enhanced version of Aristotle's syllogisms, its applicability to natural language argumentation remains limited (see our discussion above).
In contrast to our data source, a few synthetic datasets for general natural language reasoning have been recently introduced, such as answers to questions over a described physical world \cite{Weston.et.al.2016.ICLR} or an evaluation set of 100 questions in the Winograd Schema Challenge \cite{Levesque.et.al.2012}.

Finally, we note that, although being related, research on argument mining, argumentation quality, and stance classification is not in the immediate scope of this paper. For details on these, we therefore refer to recent papers from \newcite{Lippi.Torroni.2016,Habernal.Gurevych.2017.COLI} or \newcite{Mohammad.et.al.2016.SemEval}.

\section{Argument Reasoning Comprehension}
\label{sec:argument-reasoning-task}

Let $R$ be a reason for a claim $C$, both of which being propositions extracted from a natural language argument. Then there is a warrant $W$ that justifies the use of $R$ as support for $C$, but $W$ is left implicit.\,\,\,\,

For example, in a discussion about whether declawing a cat should be illegal, an author takes the following position (which is her claim $C$): `It should be illegal to declaw your cat'. She gives the following reason ($R$): `They need to use their claws for defense and instinct'.%
\footnote{The example is taken from our dataset introduced below.} 
The warrant $W$ could then be `If cat needs claws for instincts, declawing would be against nature' or similar. $W$ remains implicit, because $R$ already implies $C$ quite obviously and so, according to common sense, any further explanation seems superfluous.

Now, the question is how to find the warrant $W$ for a given reason $R$ and claim $C$. Our key hypothesis in the definition of the argument reasoning comprehension task is the existence of an \emph{alternative warrant} $AW$ that justifies the use of $R$ as support for the opposite $\neg C$ of the claim $C$ (regardless of the question of how strong this justification is).

For the example above, assume that we `twist' $C$ to `It should be \emph{legal} to declaw your cat' ($\neg C$) but use the same reason $R$. Is it possible to come up with an alternative warrant $AW$ that justifies $R$? In the given case, `most house cats don't face enemies' would bridge $R$ to $\neg C$ quite plausibly. If we now use a reasoning based on $AW$ but twist $AW$ again such that it leads to the claim $C$, we get `most house cats face enemies', which is a plausible warrant $W$ for the original argument containing $R$ and $C$.
\footnote{This way, we also reveal the weakness of the original argument that was hidden in the implicit premise. It can be challenged by asking the arguer whether house cats really face enemies.}

Constructing an alternative warrant is not possible for all reason/claim pairs; in some reasons the arguer's position is deeply embedded. As a result, trying to give a plausible reasoning for the opposite claim $\neg C$ either leads to nonsense or to a proposition that resembles a rebuttal rather than a warrant \cite{Toulmin.1958}. However, if both $W$ and $AW$ are available, they usually capture the core of a reason's relevance and reveal the implicit presuppositions (examples follow further below). 

Based on our key hypothesis, we define the argument reasoning comprehension task as: 

\emph{Given a reason $R$ and a claim $C$ along with the title and a short description of the debate they occur in, identify the correct warrant $W$ from two candidates: the correct warrant $W$ and an incorrect alternative warrant $AW$.} 

\smallskip
An instance of the task is thus basically given by a tuple $(R, C, W, AW)$. The debate title and description serve as the context of $R$ and $C$. As it is binary, we propose to evaluate the task using accuracy.

\section{Reconstruction of Implicit Warrants}
\label{sec:dataset-construction}

We now describe our methodology to systematically reconstruct implicit warrants, along with the scalable crowdsourcing process that operationalizes the methodology. The result of the process is a dataset with authentic instances $(R, C, W, AW)$ of the argument reasoning comprehension task.

\subsection{Source Data}

Instead of extending an existing dataset, we decided to create a new one from scratch, because we aimed to study a variety of controversial issues in user-generated web comments and because we sought for a dataset with a permissive license.

As a source, we opted for the \emph{Room for Debate} section of the New York Times.%
\footnote{\url{https://www.nytimes.com/roomfordebate}} 
It provides authentic argumentation on contemporary issues with good editorial work and moderation --- as opposed to debate portals such as \emph{createdebate.com}, where classroom assignments, silly topics, and bad writing prevail.
We manually selected 188 debates with polar questions in the title. These questions are controversial and provoking, giving a stimulus for stance-taking and argumentation.\footnote{Detailed theoretical research on polar and alternative questions can be found in \cite{vanRooy.Safarova.2003}; \newcite{Asher.Reese.2005} analyze bias and presupposition in polar questions.}
For each debate we created two explicit opposing claims, e.g., `It should be illegal to declaw your cat' and `It should be legal to declaw your cat'. We crawled all comments from each debate and sampled about 11k high-ranked, root-level comments.%
\footnote{To remove `noisy' candidates, we applied several criteria, such as the absence of quotations or URLs and certain lengths. For details, see the source code we provide. We did not check any quality criteria of arguments, as this was not our focus; see, e.g., \cite{Wachsmuth.et.al.2017.ACL} for argumentation quality.}

\subsection{Methodology and Crowdsourcing Process}

\begin{figure*}
\centering
\includegraphics[width=\textwidth]{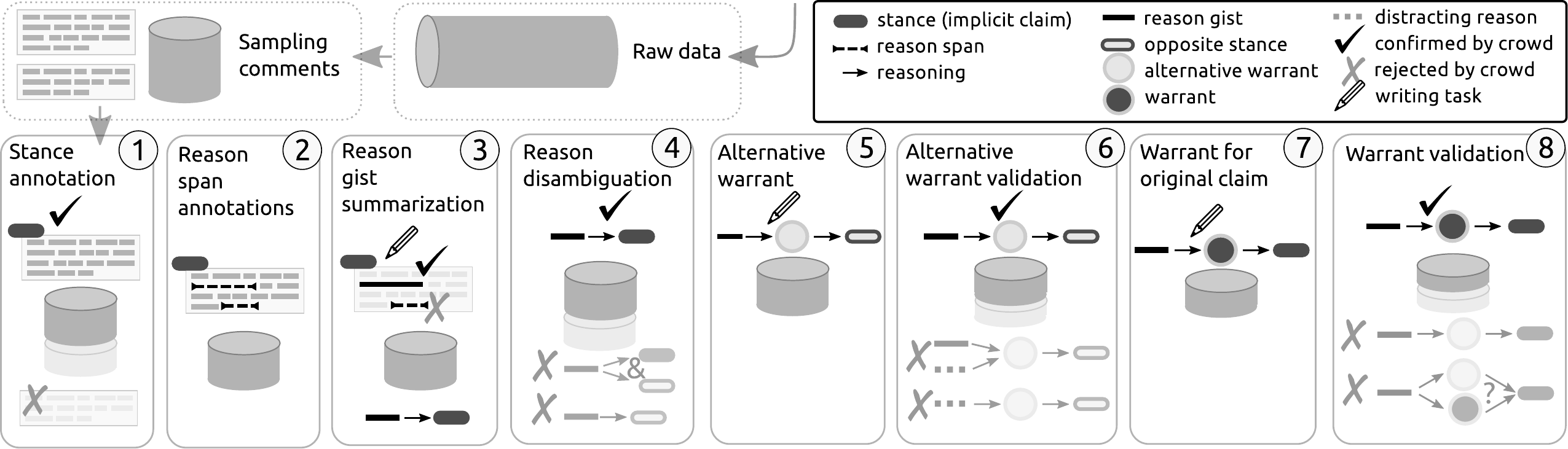}
\vspace{-2em}
\caption{\label{fig:annotation-process}Overview of the methodology of reconstructing implicit warrants for argument reasoning comprehension.}
\end{figure*}

The methodology we propose consists of eight consecutive steps that are illustrated in Figure~\ref{fig:annotation-process} and detailed below. Each step can be operationalized with crowdsourcing. For our dataset, we performed crowdsourcing on 5,000 randomly sampled comments using Amazon Mechanical Turk (AMT) from December 2016 to April 2017. Before, each comment was split into elementary discourse units (EDUs) using SistaNLP \cite{Surdeanu.2015}.

\paragraph{1.\ Stance Annotation}
For each comment, we first classify what stance it is taking (recall that we always have two explicit claims with opposing stance). Alternatively, it may be neutral (considering both sides) or may not take any stance.%
\footnote{We also experimented with approaching the annotations top-down starting by annotating explicit claims, but the results were unsatisfying. This is in line with empirical observations made by \newcite{Habernal.Gurevych.2017.COLI} who showed that the majority of claims in user-generated arguments are implicit.}

All 2,884 comments in our dataset classified as stance-taking by the crowdworkers were then also annotated as to whether being sarcastic or ironic; both pose challenges in analyzing argumentation not solved so far \cite{Habernal.Gurevych.2017.COLI}. 

\paragraph{2.\ Reason Span Annotation}
For all comments taking a stance, the next step is to select those spans that give a reason for the claim (with a single EDU as the minimal unit). 

In our dataset, the workers found 5,119 reason spans, of which 2,026 lay within arguments. About 40 comments lacked any explicit reason.

\paragraph{3.\ Reason Gist Summarization} 
This new task is, in our view, crucial for downstream annotations. Each reason from the previous step is rewritten, such that the reason's gist in the argument remains the same but the clutter is removed (examples are given in the supplementary material which is available both in the ACL Anthology and the project GitHub site). Besides, wrongly annotated reasons are removed in this step. The result is pairs of reason~$R$ and claim~$C$. 

All 4,294 gists in our dataset were summarized under Creative Commons Zero license (CC-0).

\paragraph{4.\ Reason Disambiguation} 
Within our methodology, we need to be able to identify to what extent a reason itself implies a stance: While `$C$ because $R$' allows for many plausible interpretations (as discussed above), whether $R \rightarrow C$ or $R \rightarrow \neg C$ depends on how much presupposition is encoded in $R$. In this step, we decide which claim ($C$ or $\neg C$) is most plausible for $R$, or whether both are similarly plausible (in the given data, respective reasons  turned out to be rather irrelevant though). 

We used only those 1,955 instances where $R$ indeed implied $C$ according to the workers, as this suggests at least some implicit presupposition in $R$.

\paragraph{5.\ Alternative Warrant} 
This step is the trickiest, since it requires both creativity and `brain twisting'. As exemplified in Section~\ref{sec:argument-reasoning-task}, a plausible explanation needs to be given why $R$ supports $\neg C$ (i.e., the alternative warrant $AW$). Alternatively, this may be classified as being impossible. 

Exact instructions for our workers can be found in the provided sources. All 5,342 alternative warrants in our dataset are written under CC-0 license. 

\paragraph{6.\ Alternative Warrant Validation} 
As the previous step produces largely uncontrolled writings, we validate each fabricated alternative warrant $AW$ as to whether it actually relates to the reason~$R$. To this end, we show $AW$ and $\neg C$ together with two alternatives: $R$ itself and a distracting reason. Only instances with correctly validated $R$ are kept.

For our dataset, we sampled the distracting reason from the same debate topic, using the most dissimilar to $R$ in terms of skip-thought vectors \cite{kiros2015skip} and cosine similarity. We kept 3,791 instances, for which the workers also rated how `logical' the explanation of $AW$ was (0--2 scale). 

\paragraph{7.\ Warrant For Original Claim} 
This step refers to the second task in the example from Section~\ref{sec:argument-reasoning-task}: Given $R$ and $C$, make minimal modifications to the alternative warrant $AW$, such that it becomes an actual warrant~$W$ (i.e., such that $R \rightarrow W \rightarrow C$). 

For our dataset, we restricted this step to those 2,613 instances that had a `logic score' of at least 0.68 (obtained from the annotations mentioned above), in order to filter out nonsense alternative warrants. All resulting 2,447 warrants were written by the workers again under CC0 license.

\paragraph{8.\ Warrant Validation} 
To ensure that each tuple $(R, C, W, AW)$ allows only one logical explanation (i.e., either $R \rightarrow W \rightarrow C$ or $R \rightarrow AW \rightarrow C$ is correct, not both), all instances are validated again. 

Disputed cases in the dataset (according to our workers) were fixed by an expert to ensure quality. We ended up with 1,970 instances to be used for the argument reasoning comprehension task.

\begin{table*}[t]
\begin{scriptsize}
\begin{tabularx}{\textwidth}{@{}p{0.1em}YYp{2em}>{\raggedright\arraybackslash}p{9em}p{1.9em}Y>{\raggedright\arraybackslash}p{10em}@{}}
\textbf{\#}	&\textbf{Methodology Step}	&\textbf{Input Sata}	&\textbf{Size}	&\textbf{Output Data}	&\textbf{Size}	&\textbf{Quality Assurance}	&\textbf{Use of Data}\\
\midrule
1	&Stance annotation	&Comment, topic	&5,000	&Stance-taking arguments	&2,884	&Cohen's $\kappa$ 0.58	&Argument stance detection; sarcastic argument detection \\
\midrule
2	&Reason span annotation	&Stance-taking argument	&2,884	&Reason spans (in arguments)	&5,119 (2,026)	&Krippendorff's $\alpha_{\textrm{u}}$ 0.51	&Argument component detection; argumentative text segmentation\\
\midrule
3	&Reason gist summarization	&Claim, reason span	&5,119	&Summarized reason gists (in arguments)	& 4,294 (1,927) &Qualified workers, manual inspection	& Abstractive argument summarization; reason clustering; empirical analysis of controversies\\
\midrule
4	&Reason disambiguation	&Reason gist, both claims	&4,235	&Reasons implying original stance	&1,955	&Cohen's $\kappa$ 0.42 (task-important categories)	&Argument component stance detection\\
\midrule
5	&Writing alternative warrant	&Reason gist, opposing claim	&1,955	&Fabricated warrant for reason and opposing claim	&5,342	&Qualified workers, manual inspection	&--	\\
\midrule
6	&Alternative warrant validation	&Opposing claim, alternative warrant, reason, distracting reason	&5,342	&Plausible triple of reason, alternative warrant, and opposing claim	&3,791	&
--
&Reason/Warrant relevance detection	\\
\midrule
7	&Writing warrant for original claim	&Claim, reason, alternative warrant	&2,613*	&Warrant similar to alternative warrant for reason and claim	&2,447	&Qualified workers, manual inspection	&--\\
\midrule
8	&Warrant validation	&Claim, reason, warrant, alternative warrant	&2,447	&Validated triple of reason, warrant, and claim	&1,970	&Qualified workers, experts for hard cases &Argument reasoning comprehension (our main task)\\
\bottomrule
\end{tabularx}
\end{scriptsize}
\caption{\label{tab:pipeline-overview} Details and statistics of the datasets resulting from the eight steps of our methodology implemented in a crowdsourcing process. *Input instances were filtered by their `logic score' assigned in Step 6, such that the weakest 30\% were discarded. A more detailed description is available in the readme file of the source code.}
\end{table*}

\subsection{Agreement and Dataset Statistics}

To strictly assess quality in the entire crowdsourcing process, we propose an evaluation method that enables `classic' inter-annotator agreement measures for crowdsourcing, such as Fleiss' $\kappa$ or Krippendorff's $\alpha$. Applying $\kappa$ and $\alpha$ directly to crowdsourced data has been disputed \cite{Passonneau.Carpenter.2014}. For estimating gold labels from the crowd, several models have been proposed; we rely on MACE \cite{Hovy.et.al.2013}. Given a number of noisy workers, MACE outputs best estimates, outperforming simple majority votes. At~least five workers are recommended for a crowdsourcing task, but how reliable is the output really?

We hence collected 18 assignments per item and split them into two groups (9+9) based on their submission time. We then considered each group as an independent crowdsourcing experiment and estimated gold labels using MACE for each group, thus yielding two `experts from the crowd.' Having two independent `experts' from the crowd allowed us to compute standard agreement scores. We also varied the size of the sub-sample from each group from 1 to 9 by repeated random sampling of assignments. This revealed how the score varies with respect to the crowd size per `expert'.

\begin{figure}[t!]
\includegraphics[width=\columnwidth]{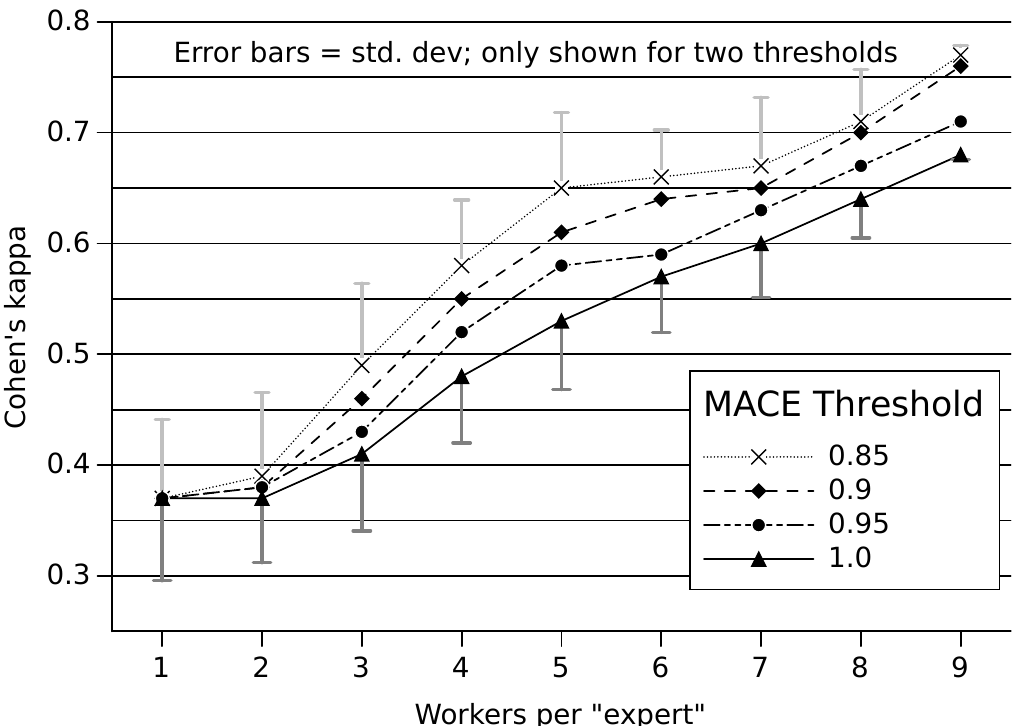}
\vspace{-2em}
\caption{\label{fig:agreement-graph-stance} Cohen's $\kappa$ agreement for stance annotation on 98 comments. As a trade-off between reducing costs (i.e., discarding fewer instances) and increasing reliability, we chose 5 annotators and a threshold of 0.95 for this task, which resulted in $\kappa$ = 0.58 (moderate to substantial agreement). }
\end{figure}

Figure \ref{fig:agreement-graph-stance} shows the Cohen's $\kappa$ agreement for stance annotation with respect to the crowd size computed by our method. As MACE also includes a threshold for keeping only the most confident predictions in order to benefit precision, we tuned this parameter, too. Deciding on the number of workers per task is a trade-off between the desired quality and the budget. For example, reason span annotation is a harder task; however, the results for six workers are comparable to those for the expert annotations of \newcite{Habernal.Gurevych.2017.COLI}.%
\footnote{The supplementary material contains a detailed figure; not to be confused with Figure \ref{fig:agreement-graph-stance} which refers to stance annotation.} 

Table \ref{tab:pipeline-overview} lists statistics of the entire crowdsourcing process carried out for our dataset, including tasks for which we created data as a by-product.

\subsection{Examples}
\label{sec:examples}

Below, we show three examples in which implicit common-sense presuppositions were revealed during the construction of the alternative warrant $AW$ and the original warrant~$W$. For brevity, we omit the debate title and description here. A full walk-through example is found in the supplementary material.

\medskip
\begin{itemize}
\small
\item[$R$:] Cooperating with Russia on terrorism ignores Russia's overall objectives.
\item[$C$:] Russia cannot be a partner.
\item[$AW$:] Russia has the same objectives of the US.
\item[$W$:] Russia has the opposite objectives of the US.\vspace{-0.75em}
\end{itemize}
\makebox[\linewidth]{\rule{\linewidth}{0.1pt}}
\begin{itemize}
\small
\item[$R$:] Economic growth needs innovation.
\item[$C$:] 3-D printing will change the world.
\item[$AW$:] There is no innovation in 3-d printing since it's unsustainable.
\item[$W$:] There is much innovation in 3-d printing and it is sustainable.\vspace{-0.75em}
\end{itemize}
\makebox[\linewidth]{\rule{\linewidth}{0.1pt}}
\begin{itemize}
\small
\item[$R$:] College students have the best chance of knowing history.
\item[$C$:] College students' votes do matter in an election.
\item[$AW$:] Knowing history doesn't mean that we will repeat it.
\item[$W$:] Knowing history means that we won't repeat it.
\end{itemize}

\section{Experiments}
\label{sec:experiments}

Given the dataset, we performed first experiments to assess the complexity of argument reasoning comprehension. To this end, we split the 1,970 instances into three sets based on the year of the debate they were taken from: 2011--2015 became the training set (1,210 instances), 2016 the development set (316 instances), and 2017 the test set (444 instances). This follows the paradigm of learning on past data and predicting on new ones. In addition, it removes much lexical and topical overlap.

\subsection{Human Upper Bounds}

To evaluate human upper bounds for the task, we sampled 100 random questions (such as those presented in Section \ref{sec:examples}) from the test set and distributed them among 173 participants of an AMT survey. Every participant had to answer 10 questions. We also asked the participants about their highest completed education (six categories) and the amount of formal training they have in reasoning, logic, or argumentation (no training, some, or extensive). In addition, they specified for each question how familiar they were with the topic (3-point scale). 

\begin{figure}[t!]
\centering
\includegraphics[width=\columnwidth]{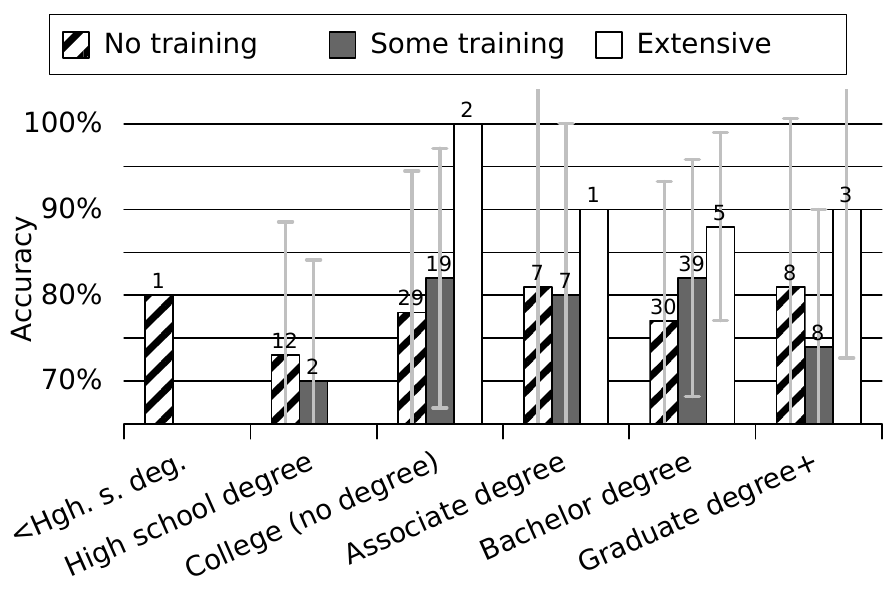}
\vspace{-2.0em}
\caption{\label{fig:human-upper-bound} Human upper bounds on the argument reasoning comprehension task with respect to education and formal training in reasoning, logic, or argumentation. For each configuration, the mean values are displayed together with the number of participants (above the bar) and with their standard deviations (error bars).}
\end{figure}

\paragraph{How Hard is the Task for Humans?} 
It depends, as shown in Figure \ref{fig:human-upper-bound}. Whereas education had almost negligible influence on the performance, the more extensive formal training in reasoning the participants had, the higher their score was. Overall, 30  of the 173 participants scored 100\%. The mean score for those with extensive formal training was 90.9\%. For all participants, the mean was 79.8\%. However, we have to note that some of the questions are more difficult than others, for which we could not control explicitly.

\paragraph{Does Topic Familiarity Affect Human Performance?}
Not really, i.e., we found no significant (Spearman) correlation between the mean score and familiarity of a participant in almost all education/training configurations. This suggests that argument reasoning comprehension skills are likely to be independent of topic-specific knowledge.

\subsection{Computational Models}

To assess the complexity of computationally approaching argument reasoning comprehension, we carried out first experiments with systems based on the following models.

The simplest considered model was the {\em random baseline}, which chooses either of the candidate warrants of an instance by chance.
As another baseline, we used a 4-gram Modified Kneser-Ney \emph{language model} trained on 500M tokens (100k vocabulary) from the C4Corpus \cite{Habernal.et.al.2016.LREC}. The effectiveness of language models was demonstrated by \newcite{Rudinger.et.al.2015.EMNLP} for the narrative cloze test where they achieved state-of-the-art results. We computed log-likelihood of the candidate warrants and picked the one with lower score.\footnote{This might seem counterintuitive, but since $W$ is created by rewriting $AW$, it may suffer from some dis-coherency, which is then caught by the language model.}

\begin{figure}[t!]
\centering
\includegraphics[width=0.9\columnwidth]{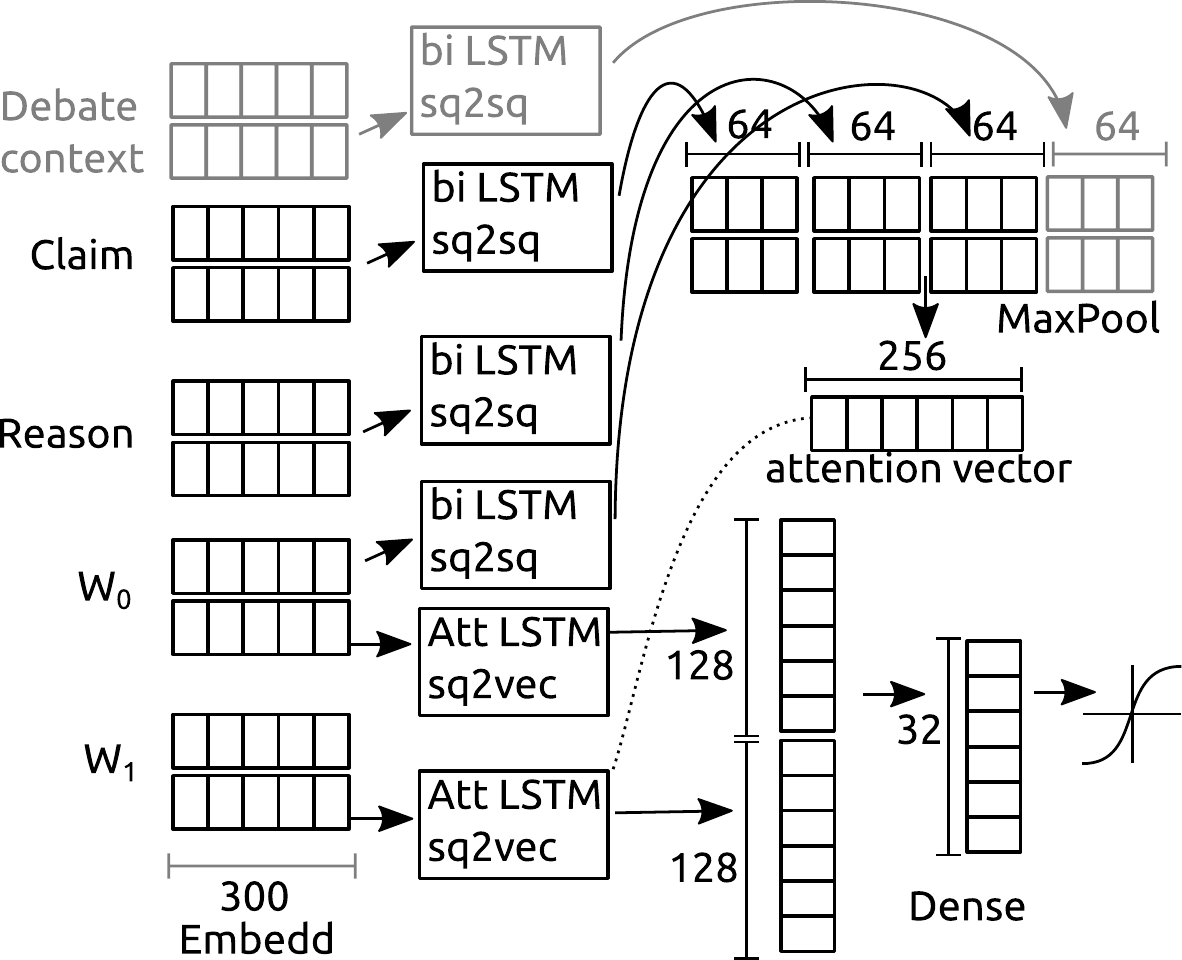}
\caption{\label{fig:network} Intra-warrant attention. Only the attention vector for the warrant $W_1$ is shown; the attention vector for $W_0$ is constructed analogously. Grey areas represent a modification with additional context.}
\end{figure}

To specifically appoach the given task, we implemented two neural models based on a bidirectional LSTM. In the standard {\em attention} version, we encoded the reason and claim using a BiLSTM and provided it as an attention vector after max-pooling to LSTM layers from the two available warrants $W_0$ and $W_1$ (corresponding to $W$ and $AW$, see below). Our more elaborated version used {\em intra-warrant attention}, as shown in Figure \ref{fig:network}. Both versions were also extended with the  debate title and description added as context to the attention layer (\emph{w/ context}). We trained the resulting four models using the ADAM optimizer, with heavy dropout (0.9) and early stopping (5 epochs), tuned on the development set. Input embeddings were pre-trained word2vec's \cite{Mikolov.2013}. We ran each model three times with random initializations.

To evaluate all systems, each instance in our dataset is represented as a tuple $(R, C, W_0, W_1)$ with a label (0 or 1). If the label is 0, $W_0$ is the correct warrant, otherwise $W_1$. Recall that we have two warrants $W$ and \emph{AW} whose correctness depends on the claim: $W$ is correct for $R$ and the original claim $C$, whereas $AW$ would be correct for $R$ and the opposite claim $\neg C$. We thus doubled the training data by adding a permuted instance $(R, C, W_1, W_0)$ with the respective correct label; this led to increased performance.
The overall results of all approaches (humans and systems) are shown in Table \ref{tab:results}. Intra-warrant attention with rich context  outperforms standard neural models with a simple attention, but it only slightly beats the language model on the dev set. The language model is basically random on the test set.

A manual error analysis of 50 random wrong predictions (a single run of the best-performing system on the dev set) revealed no explicit pattern of encountered errors. Drawing any conclusions is hard given the diversity of included topics and the variety of reasoning patterns. A possible approach would be to categorize warrants using, e.g., argumentation schemes \cite{Walton.2008} and break down errors accordingly. However, this is beyond the scope here and thus left for future work.

\paragraph{Can We Benefit from Alternative Warrants and Opposite Claims?} 
Since the reasoning chain $R \rightarrow AW \rightarrow \neg C$ is correct, too, we also tried adding respective instances to the training set (thus doubling the size). In this configuration, however, the neural models failed to learn anything better than a random guess. The reason behind is probably that the opposing claims are lexically very close, usually negated, and the models cannot pick this up.  This underlines that argument reasoning comprehension cannot be solved by simply looking at the occurring words or phrases.  

\begin{table}
\setlength{\tabcolsep}{5pt}
\begin{footnotesize}
\begin{tabularx}{\columnwidth}{@{}Yl@{\,\,\,\,}ll@{\,\,\,\,}l@{}}
\textbf{Approach}				& \textbf{Dev}	& $(\pm)$	& \textbf{Test}	& $(\pm)$ \\
\toprule
Human average				&	&	&.798	&.162\\
Human w/ training in reasoning		&	&	&.909	&.114\\
\midrule
Random baseline				&.473 & .039 & .491 & .031\\
Language model				&.617 &	& .500 &\\
\midrule
Attention						& .488 & .006 & .513 & .012 \\
Attention w/ context				&.502 & .031 & .512 & .014 \\
Intra-warrant attention			& \textbf{.638} & .024 & .556 & .016 \\
Intra-warrant attent.\ w/ context		& .637 & .040 & \textbf{.560} & .055 \\
\bottomrule
\end{tabularx}
\end{footnotesize}
\vspace{-1em}
\caption{\label{tab:results} Accuracy of each approach (humans and systems) on the development set and test set, respectively.}
\end{table}

\section{Conclusion and Outlook}

We presented a new task called \emph{argument reasoning comprehension} that tackles the core of reasoning in natural language argumentation --- implicit warrants. Moreover, we proposed a methodology to systematically reconstruct implicit warrants in eight consecutive steps. So far, we implemented the methodology in a manual crowdsourcing process, along with a strategy that enables standard~inter-annotator agreement measures in crowdsourcing.

Following the process, we constructed a new dataset with 1,970 instances for the task. This number might not seem large (e.g., compared to 0.5M from SNLI), but tasks with hand-crafted data are of a similar size (e.g., 3,744 Story Cloze Test instances). Also, the crowdsourcing process is scalable and is limited only by the budget.\footnote{In our case, the total costs were about \$6,000 including bonuses and experiments with the workflow set-up.}
Moreover, we created several data `by-products' that are valuable for argumentation research: 5,000 comments annotated with stance, which outnumbers the 4,163 tweets for stance detection of \newcite{Mohammad.et.al.2016.SemEval}; 2,026 arguments with 4,235 annotated reasons, which is six times larger than the 340 documents of \newcite{Habernal.Gurevych.2017.COLI}; and 4,235 summarized reason gists --- we are not aware of any other hand-crafted dataset for abstractive argument summarization built upon authentic arguments. 

Based on the dataset, we evaluated human performance in argument reasoning comprehension. Our findings suggest that the task is harder for people without formal argumentation training, while being solvable without knowing the topic. We also found that neural attention models outperform language models on the task.

In the short run, we plan to draw more attention to this topic by running a SemEval 2018 shared task.\footnote{\url{https://competitions.codalab.org/competitions/17327}}
A deep qualitative analysis of the warrants from the theoretical perspective of reasoning patterns or argumentation schemes is also necessary.
In the long run, an automatic generation and validation warrants can be understood as the ultimate goal in argument evaluation.
It has been claimed that for reconstructing and evaluating natural language arguments, one has to fully `roll out' their implicit premises \citep[Chap.~3.2]{vanEemeren.et.al.2014} and leverage knowledge bases \citep{Wyner.et.al.2016.ArgCompJournal}. We believe that a system that can distinguish between the wrong and the right warrant given its context will be helpful in filtering out good candidates in argument reconstruction.

For the moment, we just made a first empirical step towards exploring how much common-sense reasoning is necessary in argumentation and how much common sense there might be at all.

\section*{Acknowledgments}

This work has been supported by the ArguAna Project GU~798/20-1 (DFG), and by the DFG-funded research training group ``Adaptive Preparation of Information form Heterogeneous Sources'' (AIPHES, GRK 1994/1).

\bibliography{bibliography}

\begin{thebibliography}{}
\expandafter\ifx\csname natexlab\endcsname\relax\def\natexlab#1{#1}\fi

\bibitem[{Amossy(2009)}]{Amossy.2009}
Ruth Amossy. 2009.
\newblock \href{https://doi.org/10.1007/s10503-009-9154-y}{{The New Rhetoric's
  Inheritance. Argumentation and Discourse Analysis}}.
\newblock {\em Argumentation\/} 23(3):313--324.
\newblock \url{https://doi.org/10.1007/s10503-009-9154-y}.

\bibitem[{Angeli and Manning(2014)}]{Angeli.Manning.2014}
Gabor Angeli and Christopher~D Manning. 2014.
\newblock \href{http://www.aclweb.org/anthology/D14-1059}{{NaturalLI: Natural
  Logic Inference for Common Sense Reasoning}}.
\newblock In {\em Proceedings of the 2014 Conference on Empirical Methods in
  Natural Language Processing (EMNLP)\/}. Association for Computational
  Linguistics, Doha, Qatar, pages 534--545.
\newblock \url{http://www.aclweb.org/anthology/D14-1059}.

\bibitem[{Asher and Reese(2005)}]{Asher.Reese.2005}
Nicholas Asher and Brian Reese. 2005.
\newblock {Negative Bias in Polar Questions}.
\newblock In Emar Maier, Corien Bary, and Janneke Huitink, editors, {\em
  Proceedings of Sinn und Bedeutung (SuB9)\/}. Nijmegen Centre of Semantics,
  Nijmegen, NL, pages 30--43.

\bibitem[{Becker et~al.(2017)Becker, Staniek, Nastase, and
  Frank}]{Becker.et.al.2017.NLDB}
Maria Becker, Michael Staniek, Vivi Nastase, and Anette Frank. 2017.
\newblock \href{https://doi.org/10.1007/978-3-319-59569-6_9}{{Enriching
  Argumentative Texts with Implicit Knowledge}}.
\newblock In {\em Proceedings of NLDB\/}. Springer International Publishing,
  Li{\'{e}}ge, Belgium, pages 84--96.
\newblock \url{https://doi.org/10.1007/978-3-319-59569-6_9}.

\bibitem[{Boltu{\v{z}}i{\'{c}} and
  {\v{S}}najder(2016)}]{Boltuzic.Snajder.2016.ArgMinWS}
Filip Boltu{\v{z}}i{\'{c}} and Jan {\v{S}}najder. 2016.
\newblock {Fill the Gap! Analyzing Implicit Premises between Claims from Online
  Debates}.
\newblock In {\em Proceedings of the Third Workshop on Argument Mining\/}.
  Association for Computational Linguistics, Berlin, Germany, pages 124--133.

\bibitem[{Bowman et~al.(2015)Bowman, Angeli, Potts, and
  Manning}]{Bowman.et.al.2015}
Samuel~R. Bowman, Gabor Angeli, Christopher Potts, and Christopher~D. Manning.
  2015.
\newblock \href{http://aclweb.org/anthology/D15-1075}{A large annotated corpus
  for learning natural language inference}.
\newblock In {\em Proceedings of the 2015 Conference on Empirical Methods in
  Natural Language Processing\/}. Association for Computational Linguistics,
  Lisbon, Portugal, pages 632--642.
\newblock \url{http://aclweb.org/anthology/D15-1075}.

\bibitem[{Dagan et~al.(2009)Dagan, Dolan, Magnini, and Roth}]{Dagan.et.al.2009}
Ido Dagan, Bill Dolan, Bernardo Magnini, and Dan Roth. 2009.
\newblock \href{https://doi.org/10.1017/S1351324909990209}{Recognizing textual
  entailment: Rational, evaluation and approaches}.
\newblock {\em Natural Language Engineering\/} 15(Special Issue 04):i--xvii.
\newblock \url{https://doi.org/10.1017/S1351324909990209}.

\bibitem[{Damer(2013)}]{Damer.2013}
T.~Edward Damer. 2013.
\newblock {\em Attacking Faulty Reasoning: A Practical Guide to Fallacy-Free
  Arguments\/}.
\newblock Cengage Learning, Boston, MA, 7th edition.

\bibitem[{Feng and Hirst(2011)}]{Feng.Hirst.2011}
Vanessa~Wei Feng and Graeme Hirst. 2011.
\newblock Classifying arguments by scheme.
\newblock In {\em Proceedings of the 49th Annual Meeting of the Association for
  Computational Linguistics: Human Language Technologies - Volume 1\/}.
  Association for Computational Linguistics, Portland, Oregon, HLT '11, pages
  987--996.

\bibitem[{Freeman(2011)}]{Freeman.2011}
James~B. Freeman. 2011.
\newblock {\em Argument Structure: Representation and Theory\/}, volume~18 of
  {\em Argumentation Library\/}.
\newblock Springer Netherlands.

\bibitem[{Govier(2010)}]{Govier.2010}
Trudy Govier. 2010.
\newblock {\em A Practical Study of Argument\/}.
\newblock Wadsworth, Cengage Learning, 7th edition.

\bibitem[{Green(2014)}]{Green.2014.ArgMinWS}
Nancy~L Green. 2014.
\newblock {Towards Creation of a Corpus for Argumentation Mining the Biomedical
  Genetics Research Literature}.
\newblock In {\em Proceedings of the First Workshop on Argumentation Mining\/}.
  Association for Computational Linguistics, Baltimore, Maryland USA, pages
  11--18.

\bibitem[{Habernal and Gurevych(2017)}]{Habernal.Gurevych.2017.COLI}
Ivan Habernal and Iryna Gurevych. 2017.
\newblock \href{https://doi.org/10.1162/COLI\_a\_00276}{{Argumentation Mining
  in User-Generated Web Discourse}}.
\newblock {\em Computational Linguistics\/} 43(1):125--179.
\newblock \url{https://doi.org/10.1162/COLI\_a\_00276}.

\bibitem[{Habernal et~al.(2016)Habernal, Zayed, and
  Gurevych}]{Habernal.et.al.2016.LREC}
Ivan Habernal, Omnia Zayed, and Iryna Gurevych. 2016.
\newblock
  \href{http://www.lrec-conf.org/proceedings/lrec2016/pdf/388\_Paper.pdf}{{C4{C}orpus:
  Multilingual Web-size Corpus with Free License}}.
\newblock In {\em LREC\/}. Portoro\v{z}, Slovenia, pages 914--922.
\newblock
  \url{http://www.lrec-conf.org/proceedings/lrec2016/pdf/388\_Paper.pdf}.

\bibitem[{Hobbs et~al.(1993)Hobbs, Stickel, Appelt, and
  Martin}]{Hobbs.et.al.1993}
Jerry~R. Hobbs, Mark~E. Stickel, Douglas~E. Appelt, and Paul Martin. 1993.
\newblock \href{https://doi.org/10.1016/0004-3702(93)90015-4}{Interpretation as
  abduction}.
\newblock {\em Artificial Intelligence\/} 63(1):69 --142.
\newblock \url{https://doi.org/10.1016/0004-3702(93)90015-4}.

\bibitem[{Hovy et~al.(2013)Hovy, Berg-Kirkpatrick, Vaswani, and
  Hovy}]{Hovy.et.al.2013}
Dirk Hovy, Taylor Berg-Kirkpatrick, Ashish Vaswani, and Eduard Hovy. 2013.
\newblock \href{http://www.aclweb.org/anthology/N13-1132}{Learning whom to
  trust with {MACE}}.
\newblock In {\em Proceedings of NAACL-HLT 2013\/}. Association for
  Computational Linguistics, Atlanta, Georgia, pages 1120--1130.
\newblock \url{http://www.aclweb.org/anthology/N13-1132}.

\bibitem[{Kiros et~al.(2015)Kiros, Zhu, Salakhutdinov, Zemel, Urtasun,
  Torralba, and Fidler}]{kiros2015skip}
Ryan Kiros, Yukun Zhu, Ruslan~R Salakhutdinov, Richard Zemel, Raquel Urtasun,
  Antonio Torralba, and Sanja Fidler. 2015.
\newblock Skip-thought vectors.
\newblock In {\em Advances in neural information processing systems\/}. pages
  3294--3302.

\bibitem[{Levesque et~al.(2012)Levesque, Davis, and
  Morgenstern}]{Levesque.et.al.2012}
Hector~J. Levesque, Ernest Davis, and Leora Morgenstern. 2012.
\newblock
  \href{http://www.aaai.org/ocs/index.php/KR/KR12/paper/download/4492/4924}{{The
  Winograd Schema Challenge}}.
\newblock In {\em Proceedings of the Thirteenth International Conference on
  Principles of Knowledge Representation and Reasoning\/}. Association for the
  Advancement of Artificial Intelligence, Rome, Italy, pages 552--561.
\newblock
  \url{http://www.aaai.org/ocs/index.php/KR/KR12/paper/download/4492/4924}.

\bibitem[{Lippi and Torroni(2016)}]{Lippi.Torroni.2016}
Marco Lippi and Paolo Torroni. 2016.
\newblock \href{https://doi.org/10.1145/2850417}{Argumentation mining: State of
  the art and emerging trends}.
\newblock {\em ACM Transactions on Internet Technology\/} 16(2):10:1--10:25.
\newblock \url{https://doi.org/10.1145/2850417}.

\bibitem[{Macagno and Walton(2014)}]{Macagno.Walton.2014}
Fabrizio Macagno and Douglas Walton. 2014.
\newblock {\em Emotive Language in Argumentation\/}.
\newblock Cambridge University Press.

\bibitem[{Mikolov et~al.(2013)Mikolov, Sutskever, Chen, Corrado, and
  Dean}]{Mikolov.2013}
Tomas Mikolov, Ilya Sutskever, Kai Chen, Greg~S Corrado, and Jeff Dean. 2013.
\newblock Distributed representations of words and phrases and their
  compositionality.
\newblock In C.~J.~C. Burges, L.~Bottou, M.~Welling, Z.~Ghahramani, and K.~Q.
  Weinberger, editors, {\em Advances in Neural Information Processing Systems
  26\/}, Curran Associates, Inc., pages 3111--3119.

\bibitem[{Mohammad et~al.(2016)Mohammad, Kiritchenko, Sobhani, Zhu, and
  Cherry}]{Mohammad.et.al.2016.SemEval}
Saif Mohammad, Svetlana Kiritchenko, Parinaz Sobhani, Xiaodan Zhu, and Colin
  Cherry. 2016.
\newblock \href{http://www.aclweb.org/anthology/S16-1003}{Semeval-2016~{T}ask
  6: Detecting stance in tweets}.
\newblock In {\em Proceedings of the 10th International Workshop on Semantic
  Evaluation (SemEval-2016)\/}. Association for Computational Linguistics, San
  Diego, California, pages 31--41.
\newblock \url{http://www.aclweb.org/anthology/S16-1003}.

\bibitem[{Mostafazadeh et~al.(2016)Mostafazadeh, Chambers, He, Parikh, Batra,
  Vanderwende, Kohli, and Allen}]{Mostafazadeh.et.al.2015.NAACL}
Nasrin Mostafazadeh, Nathanael Chambers, Xiaodong He, Devi Parikh, Dhruv Batra,
  Lucy Vanderwende, Pushmeet Kohli, and James Allen. 2016.
\newblock \href{http://www.aclweb.org/anthology/N16-1098}{{A Corpus and Cloze
  Evaluation for Deeper Understanding of Commonsense Stories}}.
\newblock In {\em Proceedings of the 2016 Conference of the North American
  Chapter of the Association for Computational Linguistics: Human Language
  Technologies\/}. Association for Computational Linguistics, San Diego, CA,
  USA, pages 839--849.
\newblock \url{http://www.aclweb.org/anthology/N16-1098}.

\bibitem[{Newman and Marshall(1991)}]{Newman.1991}
S.~Newman and C.~Marshall. 1991.
\newblock {Pushing Toulmin Too Far: Learning From an Argument Representation
  Scheme}.
\newblock Technical report, Xerox Palo Alto Research Center, Palo Alto, CA.

\bibitem[{Oswald(2016)}]{Oswald.2016}
Steve Oswald. 2016.
\newblock {Commitment attribution and the reconstruction of arguments}.
\newblock In Fabio Paglieri, Laura Bonelli, and Silvia Felletti, editors, {\em
  The Psychology of Argument: Cognitive Approaches to Argumentation and
  Persuasion\/}, College Publications, pages 17--32.

\bibitem[{Passonneau and Carpenter(2014)}]{Passonneau.Carpenter.2014}
Rebecca~J. Passonneau and Bob Carpenter. 2014.
\newblock \href{http://aclweb.org/anthology/Q/Q14/Q14-1025.pdf}{{The Benefits
  of a Model of Annotation}}.
\newblock {\em Transactions of the Association for Computational Linguistics\/}
  2:311--326.
\newblock \url{http://aclweb.org/anthology/Q/Q14/Q14-1025.pdf}.

\bibitem[{Plumer(2016)}]{Plumer.2016}
Gilbert Plumer. 2016.
\newblock \href{https://doi.org/10.1007/s10503-016-9419-1}{{Presumptions,
  Assumptions, and Presuppositions of Ordinary Arguments}}.
\newblock {\em Argumentation\/} 31(3):469--484.
\newblock \url{https://doi.org/10.1007/s10503-016-9419-1}.

\bibitem[{Rajpurkar et~al.(2016)Rajpurkar, Zhang, Lopyrev, and
  Liang}]{Rajpurkar.et.al.2016.EMNLP}
Pranav Rajpurkar, Jian Zhang, Konstantin Lopyrev, and Percy Liang. 2016.
\newblock \href{https://aclweb.org/anthology/D16-1264}{{SQuAD: 100,000+
  Questions for Machine Comprehension of Text}}.
\newblock In {\em Proceedings of the 2016 Conference on Empirical Methods in
  Natural Language Processing\/}. Association for Computational Linguistics,
  Austin, Texas, pages 2383--2392.
\newblock \url{https://aclweb.org/anthology/D16-1264}.

\bibitem[{Reed and Rowe(2004)}]{Reed.2004}
Chris Reed and Glenn Rowe. 2004.
\newblock \href{https://doi.org/10.1142/S0218213004001922}{Araucaria: software
  for argument analysis, diagramming and representation}.
\newblock {\em International Journal on Artificial Intelligence Tools\/}
  13(04):961--979.
\newblock \url{https://doi.org/10.1142/S0218213004001922}.

\bibitem[{Rudinger et~al.(2015)Rudinger, Rastogi, Ferraro, and
  Van~Durme}]{Rudinger.et.al.2015.EMNLP}
Rachel Rudinger, Pushpendre Rastogi, Francis Ferraro, and Benjamin Van~Durme.
  2015.
\newblock \href{http://aclweb.org/anthology/D15-1195}{Script induction as
  language modeling}.
\newblock In {\em Proceedings of the 2015 Conference on Empirical Methods in
  Natural Language Processing\/}. Association for Computational Linguistics,
  Lisbon, Portugal, pages 1681--1686.
\newblock \url{http://aclweb.org/anthology/D15-1195}.

\bibitem[{Sugawara and Aizawa(2016)}]{Sugawara.Aizawa.2016.EMNLP-WS}
Saku Sugawara and Akiko Aizawa. 2016.
\newblock \href{http://aclweb.org/anthology/W16-6001}{{An Analysis of
  Prerequisite Skills for Reading Comprehension}}.
\newblock In {\em Proceedings of the Workshop on Uphill Battles in Language
  Processing: Scaling Early Achievements to Robust Methods\/}. Association for
  Computational Linguistics, Austin, TX, USA, pages 1--5.
\newblock \url{http://aclweb.org/anthology/W16-6001}.

\bibitem[{Surdeanu et~al.(2015)Surdeanu, Hicks, and
  Valenzuela-Escarcega}]{Surdeanu.2015}
Mihai Surdeanu, Tom Hicks, and Marco~Antonio Valenzuela-Escarcega. 2015.
\newblock \href{http://www.aclweb.org/anthology/N15-3001}{Two practical
  rhetorical structure theory parsers}.
\newblock In {\em Proceedings of the 2015 Conference of the North American
  Chapter of the Association for Computational Linguistics: Demonstrations\/}.
  Association for Computational Linguistics, Denver, Colorado, pages 1--5.
\newblock \url{http://www.aclweb.org/anthology/N15-3001}.

\bibitem[{Toulmin(1958)}]{Toulmin.1958}
Stephen~E. Toulmin. 1958.
\newblock {\em The Uses of Argument\/}.
\newblock Cambridge University Press.

\bibitem[{van Eemeren et~al.(2014)van Eemeren, Garssen, Krabbe,
  Snoeck~Henkemans, Verheij, and Wagemans}]{vanEemeren.et.al.2014}
Frans~H. van Eemeren, Bart Garssen, Erik C.~W. Krabbe, A.~Francisca
  Snoeck~Henkemans, Bart Verheij, and Jean H.~M. Wagemans. 2014.
\newblock {\em Handbook of Argumentation Theory\/}.
\newblock Springer, Berlin/Heidelberg.

\bibitem[{van Eemeren et~al.(1987)van Eemeren, Grootendorst, and
  Kruiger}]{vanEemeren.et.al.1987}
Frans~H. van Eemeren, Rob Grootendorst, and Tjark Kruiger. 1987.
\newblock {\em Handbook of argumentation theory: A critical survey of classical
  backgrounds and modern studies\/}.
\newblock Foris Publications, Dordrecht, Netherlands.

\bibitem[{van Rooy and
  {\v{S}}af{\'{a}}\v{r}ov{\'{a}}(2003)}]{vanRooy.Safarova.2003}
Robert van Rooy and Marie {\v{S}}af{\'{a}}\v{r}ov{\'{a}}. 2003.
\newblock {On polar questions}.
\newblock In Robert~B. Young and Yuping Zhou, editors, {\em Proceedings of the
  13th Semantics and Linguistic Theory Conference (SALT XIII)\/}. Ithaka, NY,
  USA, pages 292--309.

\bibitem[{Wachsmuth et~al.(2017)Wachsmuth, Naderi, Habernal, Hou, Hirst,
  Gurevych, and Stein}]{Wachsmuth.et.al.2017.ACL}
Henning Wachsmuth, Nona Naderi, Ivan Habernal, Yufang Hou, Graeme Hirst, Iryna
  Gurevych, and Benno Stein. 2017.
\newblock \href{http://aclweb.org/anthology/P17-2039}{{Argumentation Quality
  Assessment: Theory vs. Practice}}.
\newblock In {\em Proceedings of the 55th Annual Meeting of the Association for
  Computational Linguistics (Volume 2: Short Papers)\/}. Association for
  Computational Linguistics, Vancouver, Canada, pages 250--255.
\newblock \url{http://aclweb.org/anthology/P17-2039}.

\bibitem[{Walker et~al.(2014)Walker, Vazirova, and
  Sanford}]{Walker.et.al.2014.ArgMinWS}
Vern~R Walker, Karina Vazirova, and Cass Sanford. 2014.
\newblock {Annotating Patterns of Reasoning about Medical Theories of Causation
  in Vaccine Cases: Toward a Type System for Arguments}.
\newblock In {\em Proceedings of the First Workshop on Argumentation Mining\/}.
  Association for Computational Linguistics, Baltimore, Maryland USA, pages
  1--10.

\bibitem[{Walton(2007{\natexlab{a}})}]{Walton.2007}
Douglas Walton. 2007{\natexlab{a}}.
\newblock {\em Dialog Theory for Critical Argumentation\/}.
\newblock John Benjamins Publishing Company, 5 edition.

\bibitem[{Walton(2007{\natexlab{b}})}]{Walton.2007a}
Douglas Walton. 2007{\natexlab{b}}.
\newblock {\em Media Argumentation: Dialect, Persuasion and Rhetoric\/}.
\newblock Cambridge University Press.

\bibitem[{Walton(2012)}]{Walton.2012}
Douglas Walton. 2012.
\newblock Using argumentation schemes for argument extraction: A bottom-up
  method.
\newblock {\em International Journal of Cognitive Informatics and Natural
  Intelligence\/} 6(3):33--61.

\bibitem[{Walton et~al.(2008)Walton, Reed, and Macagno}]{Walton.2008}
Douglas Walton, Christopher Reed, and Fabrizio Macagno. 2008.
\newblock {\em Argumentation Schemes\/}.
\newblock Cambridge University Press.

\bibitem[{Weston et~al.(2016)Weston, Bordes, Chopra, Rush, van
  Merri{\"{e}}nboer, Joulin, and Mikolov}]{Weston.et.al.2016.ICLR}
Jason Weston, Antoine Bordes, Sumit Chopra, Alexander~M. Rush, Bart van
  Merri{\"{e}}nboer, Armand Joulin, and Tomas Mikolov. 2016.
\newblock \href{http://arxiv.org/abs/1502.05698}{{Towards AI-Complete Question
  Answering: A Set of Prerequisite Toy Tasks}}.
\newblock In {\em Procedings of the 5th International Conference on Learning
  Representations (ICLR)\/}. San Juan, Puerto Rico, pages 1--14.
\newblock \url{http://arxiv.org/abs/1502.05698}.

\bibitem[{Wilson and Sperber(2004)}]{Wilson.Sperber.2004}
Deirdre Wilson and Dan Sperber. 2004.
\newblock {Relevance Theory}.
\newblock In Laurence~R. Horn and Gregory Ward, editors, {\em The Handbook of
  Pragmatics\/}, Wiley-Blackwell, Oxford, UK, chapter~27, pages 607--632.

\bibitem[{Wyner et~al.(2016)Wyner, van Engers, and
  Hunter}]{Wyner.et.al.2016.ArgCompJournal}
Adam Wyner, Tom van Engers, and Anthony Hunter. 2016.
\newblock \href{https://doi.org/10.3233/AAC-160002}{{Working on the argument
  pipeline: Through flow issues between natural language argument, instantiated
  arguments, and argumentation frameworks}}.
\newblock {\em Argument {\&} Computation\/} 7(1):69--89.
\newblock \url{https://doi.org/10.3233/AAC-160002}.

\end{thebibliography}
\bibliographystyle{acl_natbib}

\end{document}